\documentclass[10pt,twocolumn,letterpaper]{article}

\usepackage[pagenumbers]{iccv} 

\usepackage{tikz}
\usepackage{circuitikz}
\usepackage{svg}
\usepackage{fontawesome5}
\usepackage{booktabs}
\usepackage{multirow}
\usepackage{color,soul}
\usepackage{tabularx}
\usepackage{float}
\usepackage{cuted}
\usepackage{caption}
\usepackage{graphicx}
\usepackage[accsupp]{axessibility}  

\definecolor{iccvblue}{rgb}{0.21,0.49,0.74}
\usepackage[pagebackref,breaklinks,colorlinks,allcolors=iccvblue]{hyperref}

\def\paperID{12} 
\def\confName{ICCV}
\def\confYear{2025}

\title{Sari Sandbox: A Virtual Retail Store Environment for Embodied AI Agents}

\author{Janika Deborah Gajo$^{1}$, Gerarld Paul Merales$^{1}$, Jerome Escarcha$^{1}$,\\Brenden Ashley Molina$^{1}$, Gian Nartea$^{1}$, Emmanuel G. Maminta$^{2}$,\\Juan Carlos Roldan$^{2}$, Rowel O. Atienza$^{1,2}$\\
$^{1}$EEEI, University of the Philippines, Diliman, Quezon City\\
$^{2}$AI Graduate Program, University of the Philippines, Diliman, Quezon City\\
{\tt\small jbgajo@up.edu.ph, gmmerales@up.edu.ph, jeescarcha@up.edu.ph}\\
{\tt\small bqmolina@up.edu.ph, gdnartea@up.edu.ph, emmanuel.maminta@eee.upd.edu.ph}\\
{\tt\small jtroldan@up.edu.ph, rowel@eee.upd.edu.ph}
}

\begin{document}
\maketitle

\begin{strip}
\centering
\vspace{-3em}
\includegraphics[width=\textwidth]{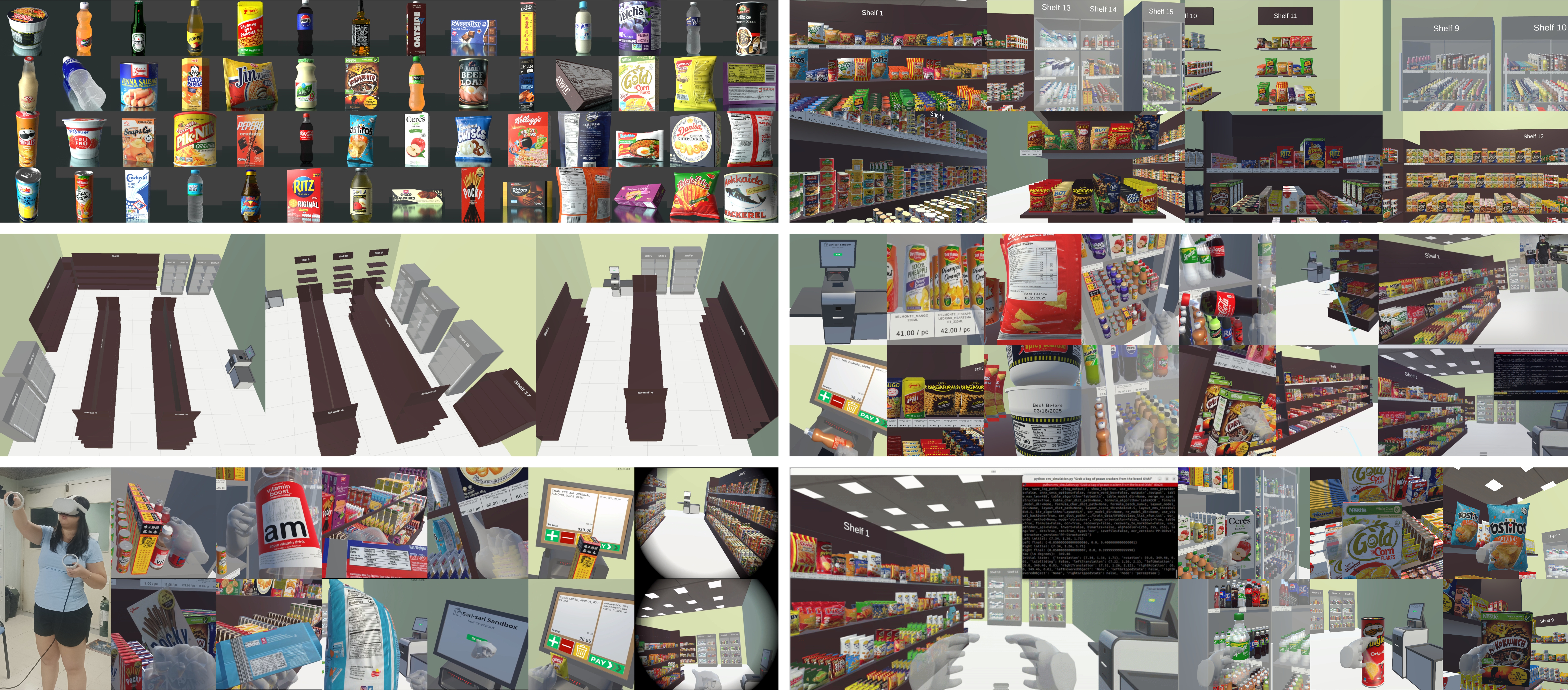}
\vspace{-1em}
\captionsetup{hypcap=false}
\captionof{figure}{Overview of our virtual retail environment for embodied AI and human benchmarking.
\textbf{Top row}: Photorealistic, high-fidelity 3D product models (left) and their randomized placement within semantically grouped categories (right).
\textbf{Middle row}: Three store layouts and key features such as a functional checkout, dynamic labels (price, expiration), interactable products and elements, teleportation system, VR support, and a Python API for agents.
\textbf{Bottom row}: Human participants in VR (left) perform shopping tasks (e.g., picking, inspecting, checkout, navigation), with optional tunneling vignette. An embodied agent (right) completes comparable tasks for benchmarking.}
\label{fig:teaser}
\end{strip}

\begin{abstract}

We present \textbf{Sari Sandbox}, a high-fidelity, photorealistic 3D retail store simulation for benchmarking embodied agents against human performance in shopping tasks. Addressing a gap in retail-specific sim environments for embodied agent training, Sari Sandbox features over 250 interactive grocery items across three store configurations, controlled via an API. It supports both virtual reality (VR) for human interaction and a vision language model (VLM)-powered embodied agent. We also introduce \textbf{SariBench}, a dataset of annotated human demonstrations across varied task difficulties. Our sandbox enables embodied agents to navigate, inspect, and manipulate retail items, providing baselines against human performance. We conclude with benchmarks, performance analysis, and recommendations for enhancing realism and scalability. The source code can be accessed via {\small \texttt{https://github.com/upeee/sari-sandbox-env}.}
\end{abstract}
\usetikzlibrary{shapes.geometric,shapes.symbols,fit,positioning,shadows}
\section{Introduction}
\label{sec:intro}

The demand for understanding retail behaviors is rapidly evolving. Paolanti et al.~\cite{paolanti2020shopper_behavior} developed an RGB-D-based deep learning system for analyzing shopper movement, shelf interactions, and re-identification, allowing store-level behavioral insights. While such systems are valuable for observing real human behavior, there is a growing interest in studying these phenomena within photorealistic simulation environments that support automated agents. These environments enables us to systematically test hypotheses about product selection, navigation, and layout design---on-scale and under controlled conditions. For example, NVIDIA's Omniverse is a powerful platform used by popular retailers like Kroger~\cite{kroger_nvidia2022} and Lowe's~\cite{lowes_storedigitaltwin2022} to construct digital twins of their respective stores. Simulation environments are a great way to enable scalable, cost-effective and realistic evaluations of agents performing these complex tasks~\cite{afzal2020challenges_simulators, kargar2024emerging_simulations}.

Specialist embodied agents in retail settings are trained in these specialized simulation platforms. In Lowe's case, they created an interactive Omniverse replica of a home improvement store, where retail employees operate augmented reality (AR) headsets to overlay the digital twin in the physical store. Several similar ideas about robotics and digital twin systems proliferate in the Future Convenience Store Challenge (FCSC) at the World Robot Summit~\cite{WRSFCSC2024}. FCSC proposes tasks for the acquisition and placement of items in a custom retail store layout. To this end, the organizers of this event have even exposed Gazebo/ROS packages for the virtual store layout and CAD-based models for these tasks.

Indoor navigation simulators are mainly focused on domestic environments, which typically refer to household settings such as kitchens, bedrooms, and living rooms~\cite{wong2025survey}, as seen in Habitat~\cite{puig2023habitat3}, AI2Thor~\cite{kolve2017ai2-thor}, Matterport3D~\cite{chang2017matterport3d} and ThreeDWorld~\cite{gan2021threedworld}. Little work is done on embodied agent simulation for retail activities, product and shelf stocking, and grocery store layout. We note that the tasks exposed by the FCSC \cite{wrs2025_fcsc} include restocking of items and store renovation. In this work, we propose exploring sim environments modeling tasks such as more complex item selection for checkout and product comparison with a more diverse product set. To accomplish these, we introduce the following contributions.
\begin{itemize}
    \item \textbf{Sari Sandbox} is a virtual retail environment for developing and evaluating embodied agents. Modeled after small convenience stores, it features interactive 3D layouts, photorealistic rendering of 250 diverse products, and three distinct store layouts based on surveys of real local configurations.
    \item We provide a Python API for agent control and data collection within the Sari Sandbox environment.
    \item We publish \textbf{SariBench}, a benchmark of retail tasks performed in the Sari Sandbox environment plus human demonstrations of these tasks using VR interaction with the Sari Sandbox, as a baseline for testing embodied agent performance on the aforementioned tasks.
    \item Finally, we design an agentic AI architecture to address the easy tasks in the SariBench. This embodied agent leverages off-the-shelf proprietary and open-source models, employing a straightforward architecture and toolset without requiring any fine-tuning. The discussion can be found in the Supplementary Material.
\end{itemize}

The next sections cover related virtual store simulators and agent designs, detail the Sari Sandbox environment's design and benchmarks, present performance profiling and comparisons between humans and agents, and conclude with recommendations for future work. An overview of the environment is shown in Figure \ref{fig:teaser}.


\section{Review of related work}
\subsection{Embodied retail store simulators}

\begin{table}[ht]
\begin{center}
\caption{Comparison of visual navigation datasets using RGB sensor data. Notably, these benchmarks primarily focus on domains other than retail, highlighting a research gap (Room-to-Room: household navigation; ION \& ALFRED: household tasks; HumanoidBench: humanoid locomotion and manipulation).}
\label{tab:nav_datasets}
\small
\begin{tabular}{lll} 
\toprule
\textbf{Benchmark dataset} & \textbf{Size} & \textbf{Simulator}\\
\hline
Room-to-Room~\cite{ku2020room} & 90 scenes & Matterport3D~\cite{chang2017matterport3d}\\
ION~\cite{li2021ion} & 600 scenes & AI2-THOR~\cite{kolve2017ai2-thor}\\
HumanoidBench~\cite{sferrazza2024humanoidbench} & 27 tasks & MuJoCo~\cite{mujoco2012todorov}\\
ALFRED~\cite{shridhar2020alfred} & 120 scenes & AI2-THOR~\cite{kolve2017ai2-thor}\\
\bottomrule
\end{tabular}
\end{center}
\end{table}

Embodied agent simulators provide an avenue for AI to be trained at low cost and with minimal risks, while still allowing for complex, interactive learning experiences. The primary goal of our work is to establish an environment that exposes high-fidelity, physically accurate, and photorealistic interactions. Several simulators like Meta's Habitat-Sim~\cite{puig2023habitat3}, Matterport3D-Simulator~\cite{chang2017matterport3d}, Isaac Sim~\cite{b8} powered by NVIDIA Omniverse~\cite{omniverse}, iGibson~\cite{shen2021igibson1, li2021igibson2}, and AI2Thor~\cite{kolve2017ai2-thor} have been widely used to build benchmarks for visual exploration and diverse embodied tasks in virtual space. As detailed in Table~\ref{tab:nav_datasets}, these prominent benchmark datasets leverage such simulators to focus on contexts like household navigation, general manipulation, or locomotion. A survey on sim environments~\cite{wong2025survey} outlines how such environments focusing on goal-driven navigation have slowly progressed from basic navigation to tackling complex interactions, predominantly in these non-retail settings. Another survey on embodied agents~\cite{survey_on_embodied_ai2022duan} evaluated simulators using a set of features that serve as robust evaluation criteria, including environment type, object complexity, physics fidelity, and interactivity. We adopted these criteria as a basis for Sari Sandbox's design, explicitly addressing the underexplored domain of retail scenarios.

As outlined also from that survey~\cite{survey_on_embodied_ai2022duan}, there are three research tasks that provide a foundation for embodied agent adaptation in unfamiliar environments. These are: visual exploration, wherein the embodied agent gathers data about its environment through motion and perception; visual navigation refers to embodied agent navigation usually to achieve a goal; and embodied question answering, wherein the embodied agent would be required to navigate and answer questions. We expect to establish entry points from our research into the three tasks laid above.

While existing embodied agent simulators are increasingly sophisticated, they are not tailored for the operational needs of retail. The prevalent focus on household or industrial settings limits their applicability for retail-specific tasks like inventory stocking, product recognition, or dynamic shelf management. To address this gap, we introduce Sari Sandbox, an environment designed specifically for developing and evaluating embodied agents in a retail context.

\subsection{Embodied agents in virtual simulation}

Aligning linguistic input with agent capabilities is crucial for grounding language in embodied simulations. However, a persistent challenge is bridging the gap between a language model's semantic output and the spatial precision needed for navigation. Geometric maps inherently offer more structure for this than language-only observations, as highlighted by Huang \textit{et al.} (2023)~\cite{huang23vlmaps}. Our work, similar to LM-Nav~\cite{shah2022lmnav}, explores language-guided navigation using off-the-shelf models. While LM-Nav targets physical agents and our research focuses on agentic patterns in virtual environments, both approaches grapple with translating high-level language into precise, executable actions. What distinguishes our work is the incorporation of visual inputs alongside language.

The development of LLM-powered embodied agents has shown immense potential in sandbox environments, leveraging advanced reasoning and language abilities for open-ended tasks~\cite{liu2025advances}. Projects such as NVIDIA's \textsc{Voyager}~\cite{wang2023voyager} and the proposed \textsc{Embodied Agent Interface}~\cite{li2024embodied} provide further validation of their effectiveness. A common framework adopted for such agents is ReAct (Reason and Act)~\cite{yao2023react}, whose core loop (\textit{thought} → \textit{action} → \textit{observation} → \textit{refinement}) effectively emulates human problem-solving and is highly suitable for embodied tasks. The adaptability of ReAct has been demonstrated in other work, such as \textsc{StateAct}, which extends it for the \textsc{ALFWorld} household environment~\cite{rozanov2024stateact, shridhar2020alfworld}. A critical challenge for these LLM- or VLM-powered embodied agents is their stateless nature, where each inference is an isolated event, unlike human cognition which constantly integrates memory. To overcome this, various approaches for explicit memory modeling have been proposed, including cognitive architectures like CoALA~\cite{sumers2023cognitive}, which models procedural, working, semantic, and episodic memory components. Furthermore, memory writing operations~\cite{zhang2024survey} are commonly used to manage and utilize these memories, often involving techniques like creating concise summaries of immediate surroundings~\cite{zhong2024memorybank} or informative recollections~\cite{modarressi2023ret} for grounding the agent in its environment.
\section{Design}
\label{sec:design}

Sari Sandbox is a 3D retail environment designed for evaluating embodied agents in retail tasks. It features an API-controlled avatar, three store layouts, 250 grocery products, and a self-checkout system. Each product is annotated with rich ground truth data which includes category, name, price, net weight, ingredients, nutritional facts, allergens, and manufacturing origin that could enable precise evaluation and data-driven experimentation. The environment is performance-optimized without sacrificing texture fidelity, built using Unity’s Universal Render Pipeline (URP) for its balance of rendering quality, efficiency, and decal support (crucial for dynamic on-product labels). Real-time physics and interaction are handled via Unity’s default 3D engine, NVIDIA PhysX.

\subsection{Design requirements}\label{section-3}
In order to create the environment, the following design requirements were taken into consideration:

\begin{itemize}
    \item \textbf{Environment performance}. 
    The environment should perform at 60 frames-per-second well without compromising its visual quality.
    
    \item \textbf{Diverse products}.
    The product diversity in the environment should closely resemble that of its real-world counterpart. The textures used should be high-fidelity to allow humans and AI to read the smallest text.

    \item \textbf{API}.
    Users and/or embodied agents should be able to control the avatar and the environment. In addition, they shall also receive environment data such as avatar position, avatar rotation, hand grip state, store layout, and randomization seed.

    \item \textbf{VR capabilities}.
    The environment should be connected and optimized for VR capabilities for human benchmarking with minimal user fatigue and nausea. 
    
\end{itemize}

\subsection{Models}

A total of 250 3D products were created for the environment. The products are based on real-life packaged goods typically found in a convenience store. This number was determined through an initial survey of several convenience stores, from which we compiled a representative and diverse set of food products commonly encountered by consumers. The selection was further limited to food products with packaging that could be flattened for scanning using a flatbed scanner. We then used open-source tools, namely GIMP for preprocessing and Blender for 3D modeling.

Product meshes are classified into three types: simple, complex, and deformable. Simple products (e.g., boxes) use basic primitives; complex products (e.g., bottles) combine multiple primitives; and deformable items (e.g., chip bags, juice packs) require detailed modeling due to their non-rigid structure. All products are simulated as rigid bodies. Additionally, they are categorized into 11 food types (Table~\ref{product-count}).

\begin{table}[ht]
\centering
\caption{Product count per category, with a total of 250 products across 11 categories.}
\label{product-count}
\small
\renewcommand{\arraystretch}{1.2}
\begin{tabular}{@{}ll ll ll ll@{}}
\toprule
\textbf{Water} & 12 & \textbf{Soda} & 23 & \textbf{Juice} & 16 & \textbf{Dairy} & 20 \\
\textbf{Biscuit} & 50 & \textbf{Can} & 59 & \textbf{Chips} & 40 & \textbf{Nuts} & 15 \\
\textbf{Soup} & 6 & \textbf{Noodles} & 7 & \textbf{Liquor} & 2 & & \\
\bottomrule
\end{tabular}

\end{table}

\begin{figure}
    \centering
    \includegraphics[width=0.7\linewidth]{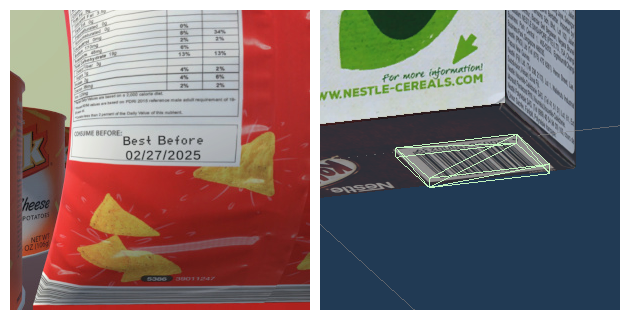}    
    \vspace{0.1em}
    \parbox[b]{0.35\linewidth}{\raggedleft (A)}%
    \hfill%
    \parbox[b]{0.35\linewidth}{\raggedright (B)}   
    \caption{Expiration decal (A) projects the generated date onto the product surface, while the barcode plane (B) is positioned above for scanner detection.}
    \label{fig:exp-date-and-barcode}
\end{figure}

\begin{figure}[htbp]
\centerline{\includegraphics[width = .20\textwidth]{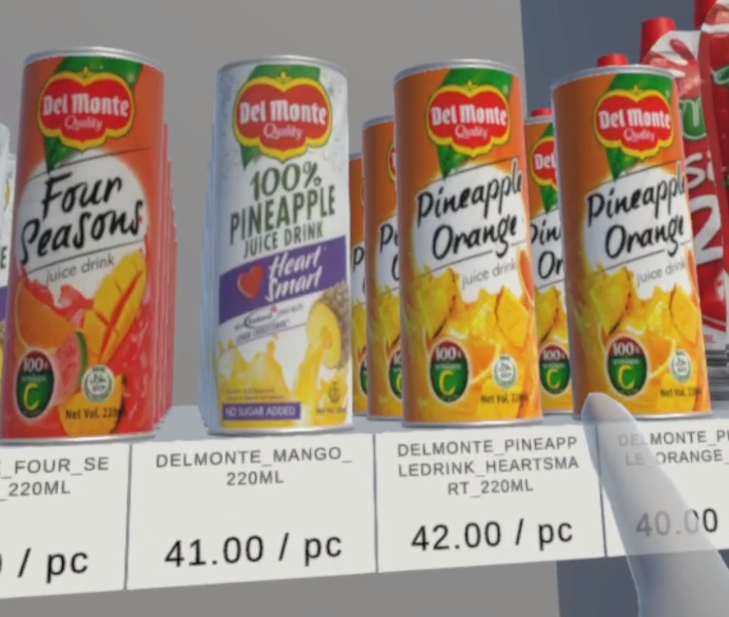}}
\caption{Price tags displayed on the front edge of the shelf.}
\label{priceTags}
\end{figure}

As Figure~\ref{fig:exp-date-and-barcode} shows, barcodes and expiration date stamps are integrated into Unity prefabs. These elements activate dynamically when the avatar grabs a product and hide otherwise, optimizing rendering and simulating realistic handling—much like how shoppers inspect packages for dates. Instead of scanning the barcode texture directly, a simulated object detection method using ray casting is employed. Rays target specific barcode planes in front of the actual barcode surface; the plane's rotation confirms correct scanning orientation. To keep expiration dates current and adaptable, each product's date is randomly generated at runtime. It is then rendered using Unity's URP decal system, which allows it to conform to various mesh shapes and update procedurally at the start of each session. Price tags, displayed along the front edge of each shelf (Figure~\ref{priceTags}), dynamically update to reflect the randomly placed products. While placements vary, the system groups similar product types, mirroring real-world grocery arrangements, with aligned price tags.

\begin{figure}
    \centering
    \includegraphics[width=0.8\linewidth]{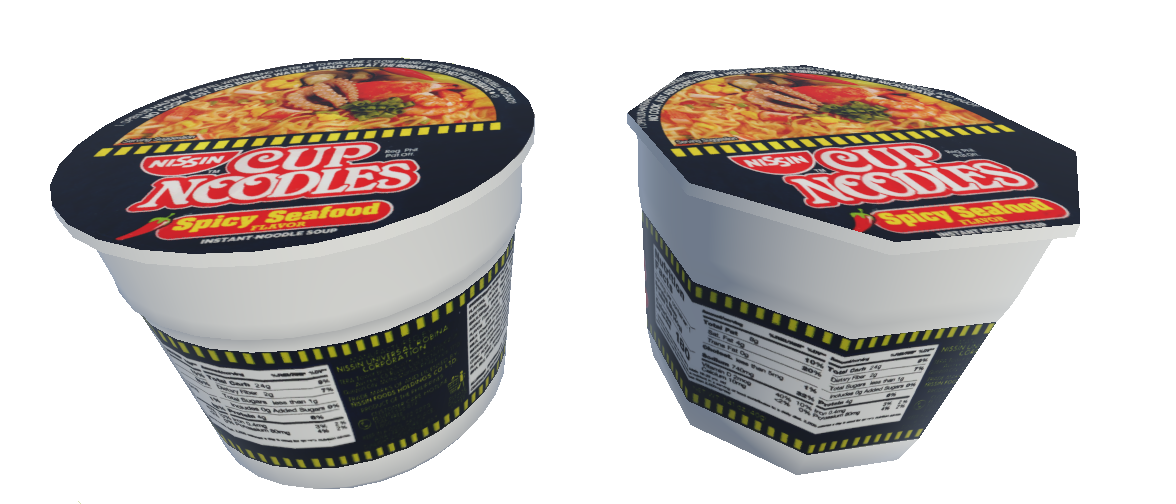}
    \vspace{0.1em}
    \parbox[b]{0.35\linewidth}{\raggedleft (A)}%
    \hfill%
    \parbox[b]{0.35\linewidth}{\raggedright (B)} 
    \caption{Level of Detail (LOD) comparison between high (A) versus low (B) quality.}
    \label{fig:level-of-detail}
\end{figure}

To enhance space and time efficiency, 3D models underwent optimization. Due to the scene's numerous objects, only box colliders are used to minimize physics computation time. High-resolution image textures, a common source of excessive space consumption, were addressed by using JPG format. We also implemented Level of Detail (LOD), a standard game development technique where objects render with varying detail based on camera distance (Figure~\ref{fig:level-of-detail}). These optimized models are then arranged within the environment to simulate real-world grocery store product displays.

\subsection{Environment}
\label{sec:environment}

\begin{figure}[htbp]
    \centerline{\includegraphics[width = .475\textwidth]{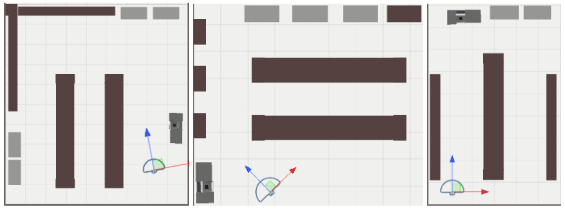}}
    \vspace{0.1em}
    \parbox[b]{0.20\linewidth}{\raggedleft (A)}%
    \hfill%
    \parbox[b]{0.15\linewidth}{\raggedleft (B)}%
    \hfill%
    \parbox[b]{0.14\linewidth}{\raggedright (C)}%
    \caption{The three store layouts: Store 1 (A), Store 2 (B), and Store 3 (C), with the avatar's starting position indicated.}
    \label{fig:store-layouts}
\end{figure}

Several real-world small-scale retail stores were surveyed to inform the design of the virtual environment, focusing on store size, arrangement, and layout. From these observations, three distinct store layouts were recreated in the environment, each featuring shelves and a self-checkout counter. The corresponding top-down view for each store layout is shown in Figure~\ref{fig:store-layouts}. Shelves are equipped with overhead labels and are assigned either single or multiple product categories, each with corresponding price tags (see Figure \ref{fig:teaser}). In addition to standard shelving, hinge door cabinets and sliding door cabinets were implemented to simulate real-world refrigeration units.

\begin{figure}[htbp]
    \centerline{\includegraphics[width = .45\textwidth]{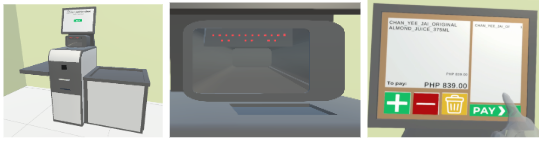}}
    \vspace{0.1em}
    \parbox[b]{0.20\linewidth}{\raggedleft (A)}%
    \hfill%
    \parbox[b]{0.06\linewidth}{\raggedleft (B)}%
    \hfill%
    \parbox[b]{0.21\linewidth}{\raggedright (C)}%
    \caption{Depicted are the self-checkout counter (A), barcode scanner (B), and touchscreen (C).}
    \label{checkout-counter}
\end{figure}

The self-checkout counter as shown in Figure \ref{checkout-counter} incorporates a touchscreen interface operable using the tip of the avatar's index finger. This interface allows users to view, add, and remove products from their virtual shopping cart. Dedicated buttons initiate the checkout process or modify scanned items. A fixed barcode scanner on the counter simulates barcode reading by casting a ray toward the barcode plane of a product, which is only successfully registered if the orientation aligns correctly with the scanner.

\begin{figure}[htbp]
    \centerline{\includegraphics[width = .4\textwidth]{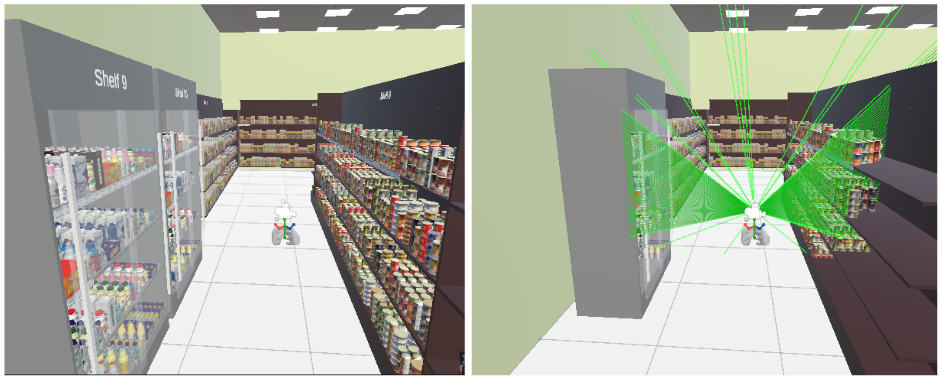}}
    \vspace{0.1em}
    \parbox[b]{0.33\linewidth}{\raggedleft (A)}%
    \hfill%
    \parbox[b]{0.31\linewidth}{\raggedright (B)} 
    \caption{Frustum and occlusion culling visualization: before (A) and after (B) application.}
    \label{fig:culling}
\end{figure}

To maintain performance in real-time rendering, frustum culling and occlusion culling were implemented. These techniques ensure that only visible objects within the camera's view are rendered (see Figure~\ref{fig:culling}), significantly improving runtime efficiency in densely packed environments.

To enhance the immersive experience and mitigate common usability and comfort issues in VR, we implemented several features based on established interaction principles~\cite{vrex,calandra2022cybersickness}. These support naturalistic interaction, user comfort, and consistency across experimental setups. Specifically, we developed a \textbf{hand interaction system} for grabbing, touching, placing, and throwing objects. We also incorporated \textbf{haptic feedback} that triggers when users hover over interactable objects, reinforcing interaction and presence. To reduce cybersickness, a \textbf{tunneling vignette} limits peripheral vision during movement. Finally, a \textbf{teleportation system} with a parabolic reticle was included to support spatial navigation while addressing physical room-scale and tracking limitations.

\subsection{Avatar}

\begin{figure}[htbp]
    \centering
    \includegraphics[width=0.4\textwidth]{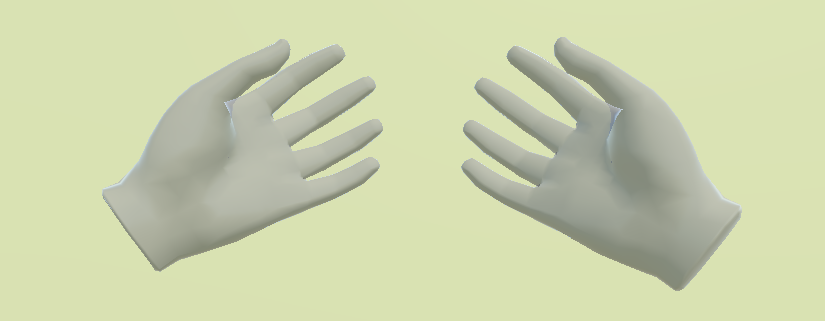}
    \caption{Hand models for the avatar.}
    \label{fig:hands}
\end{figure}

The user's in-environment representation is an avatar, comprising a camera and two hands within a character controller. For avatar interaction, we adapted the hand models shown in Figure~\ref{fig:hands}. This configuration supports interaction via either VR hardware or API-driven simulation. If no VR headset is detected at startup, the camera is positioned at an approximate adult human eye height of 1.6 meters, with hands offset 0.5 meters downward to reflect a natural standing posture.

\subsection{SariBench}
\label{sec:dataset}
Since this environment is specifically designed for retail store tasks and to the best of our knowledge, there is no existing studies have established a baseline for such scenarios, we propose a set of tasks to serve as a benchmark for evaluating embodied agent performance. These tasks and their corresponding levels are detailed in Table~\ref{baseline-task-levels}.

\begin{table}[!htbp]
\small
\begin{center}
\caption{\textbf{SariBench} tasks: Baseline tasks of varying difficulty along with the skills involved to execute them.}
\label{baseline-task-levels}
\begin{tabular}{{ p{1.25cm} p{2.5cm} p{2.25cm} }}
\toprule
\textbf{Difficulty} & \textbf{Skills involved} & \textbf{Example} \\
\hline
Easy & Perception, Navigation, Manipulation & Find and pick up a box of \textbf{cereal}. \\
Average & Perception, Navigation, Manipulation, Memory, Task Execution & Pick up a bottle of \textbf{soda} and scan at checkout.\\
Difficult & Perception, Navigation, Manipulation, Memory, Task Execution, Decision Making, Comprehension & Which of these two products has lower sugar content: \textbf{strawberry-flavored biscuit} or \textbf{chocolate-flavored biscuit}? Scan the answer. \\
\bottomrule
\end{tabular}
\end{center}
\end{table}

\begin{figure}[htbp]
    \centering
    \includegraphics[width=0.4\textwidth]{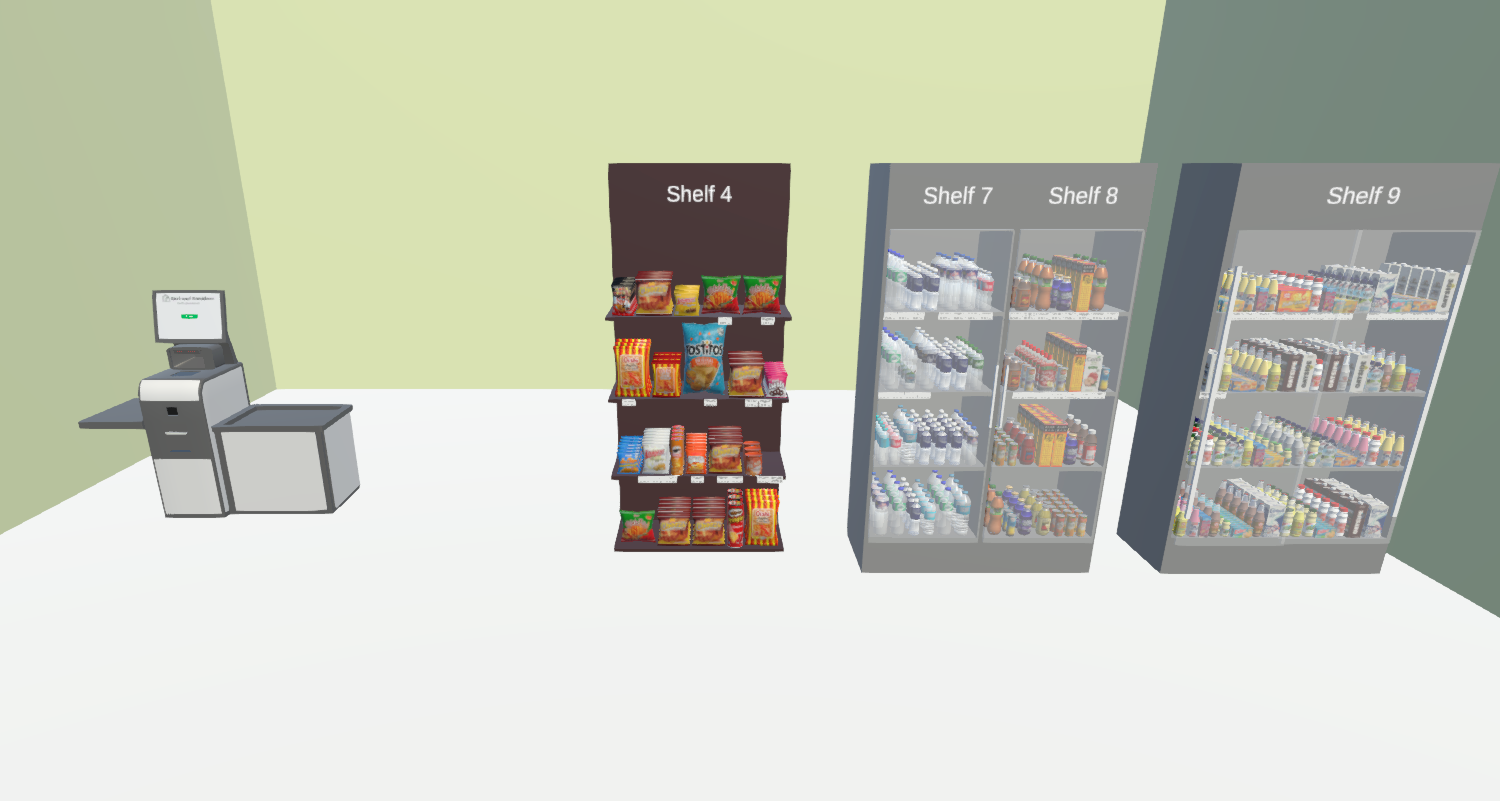}
    \caption{The playground environment used to introduce the participants to the VR controls.}
    \label{fig:playground}
\end{figure}

\begin{figure}[htbp]
\centerline{\includegraphics[width = .45\textwidth]{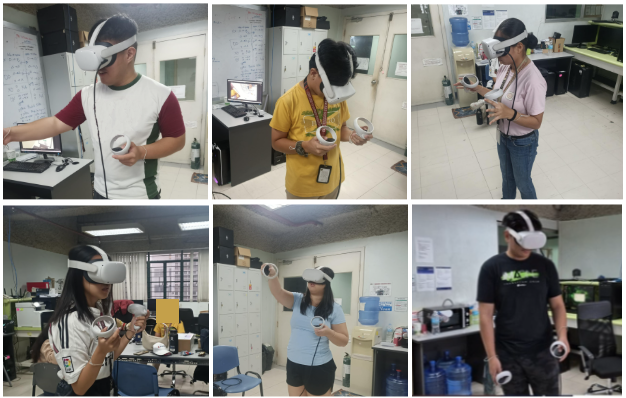}}
\caption{Volunteers testing the Sari Sandbox.}
\label{vr-benchmarking-participants-1}
\end{figure}

To create the SariBench dataset, we adapted the environment for VR headsets and recruited volunteers to perform retail tasks. The dataset currently comprises 100 videos of participants completing these tasks. Besides screen recordings, we captured the following environmental data at 10 frames per second: \textbf{global head position and rotation}, \textbf{global hand position and rotation}, \textbf{grip state}, and \textbf{hovered or held item}. Twenty human participants tested the environment, completing tasks selected from a curated pool categorized by difficulty. Each participant received a briefing on the data collection's purpose, project background, and specific data types collected. Participants had 15 minutes to familiarize themselves with VR controls in a playground environment shown in Figure~\ref{fig:playground} to ensure that they could navigate, manipulate objects, and feel comfortable in the virtual space, following established best practices for VR immersion~\cite{vrex}. Once acclimated, each participant was assigned three easy, two average, and two hard tasks. During task execution, participants were encouraged to verbalize their thought processes, which were recorded and later transcribed. Figure~\ref{vr-benchmarking-participants-1} shows several participants using the VR system to complete tasks for dataset collection.

\subsection{API}\label{subsection:api}
\begin{figure}[ht]
\centering
\scalebox{0.70}{
\begin{tikzpicture}[font=\sffamily\footnotesize]

\node[inner sep=0pt, outer sep=10pt] (user) {\includegraphics[width=1cm]{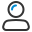}};
\node[draw, minimum width=2.5cm, minimum height=1cm, right=0cm of user, xshift=-1mm, align=center] (clientapi) {Client-side\\API};

\node[draw, minimum width=2.5cm, minimum height=1cm, right=of clientapi, xshift=2cm, align=center] (serverapi) {Server-side\\API};

\node[inner sep=0pt, outer sep=10pt, right=0cm of serverapi, xshift=-1mm] (server) {\includegraphics[width=1.5cm]{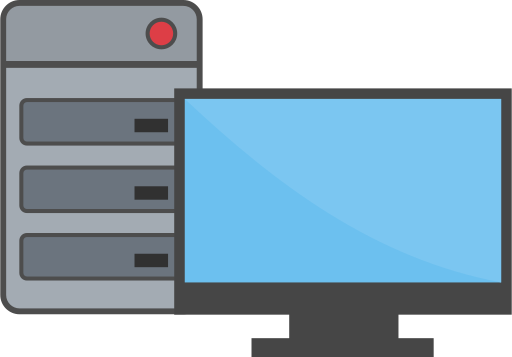}};

\draw[<->, thick] (clientapi.east) -- node[above, font=\small, align=center]{WebSocket Server} (serverapi.west);
\node[below=1pt of server, font=\sffamily\scriptsize] {Sari Sandbox};

\end{tikzpicture}
}
\caption{Communication flow between client-side and server-side APIs via WebSocket.}
\label{fig:api-diagram}
\end{figure}

The avatar and environment are controlled via a direct Python API, designed for seamless integration with an embodied agent. This API comprises a Python-based client that sends JSON files containing functions for avatar control, environment data retrieval, and simulation resets. These JSON files are received and parsed by a C\# (Unity's scripting language) server, which then executes the corresponding actions within the Unity environment. Communication between the client and server is facilitated through a WebSocket server, as illustrated in Figure~\ref{fig:api-diagram}. The API exposes three primary function types: \textit{Agent Actions}, \textit{Information Gathering}, and \textit{Store Manipulation}. Table~\ref{tab:env_api} lists all available functions; translation (T) and rotation (R) parameters accept 3D vector inputs.

\begin{table}[htbp]
\centering
\caption{Agent control API functions with full arguments. \textbf{T}: translation. \textbf{R}: rotation.}
\label{tab:env_api}
\small
\begin{tabular}{@{}p{4.3cm} p{3.6cm}@{}}
\hline
\textbf{Function} & \textbf{Description} \\
\hline
\texttt{TransformAgent(T, R)} & Manipulates the agent’s body or camera. \\
\texttt{TransformHands(leftT, leftR, rightT, rightR)} & Transforms the left and right hands. \\
\texttt{ToggleLeftGrip()} & Toggles the left hand grip to grab objects. \\
\texttt{ToggleRightGrip()} & Toggles the right hand grip to grab objects. \\
\texttt{ToggleLeftPoke()} & Toggles the left hand’s poke animation. \\
\texttt{ToggleRightPoke()} & Toggles the right hand’s poke animation. \\
\texttt{RequestScreenshot()} & Captures the current camera view. \\
\texttt{Reset()} & Resets the environment to its initial state. \\
\hline
\end{tabular}

\end{table}
\section{Experiments and analysis}

The environment was developed and executed on a desktop equipped with an Intel i7-8700 3.2GHz CPU, NVIDIA GTX 1080 GPU, and 64 GB of RAM. A Meta Quest 2 VR headset was used for dataset collection. All results and analyses presented in this paper were computed based on these hardware specifications. For the embodied agent evaluation, we used a paid API version of Gemini 2.5 Pro (\texttt{gemini-2.5-pro-preview-05-06}) with reasoning budget of 2048 tokens. The discussion about the design and development process of the basic embodied agent is found in the Supplementary Material.

\subsection{Performance metrics}\label{performance}
Using the Unity Profiler, we measured the frame processing time over 300 frames. Average frames-per-second were 26.73 (Layout 1), 23.23 (Layout), and 35.14 (Layout 3). Layout 3 exhibited the highest performance, followed by Layout 1, then Layout 2. This performance variation stems from differing store sizes, which impact the number of objects and, consequently, physics calculations during object instantiation at startup, causing a processing time spike.

\subsection{Texture fidelity}\label{texture}
\label{sec:texture-fidelity}

To assess texture fidelity, we utilized PaddleOCR~\cite{paddleocr2020}, an open-source model recognized for its robust performance in multilingual and complex text recognition~\cite{OCR-analysis-1, OCR-analysis-2}. We evaluated PaddleOCR on product label text using precision, recall, and character error rate (CER), achieving high accuracy due to the structured layout: Precision: 0.986, Recall: 0.943, CER: 0.014. However, OCR performance typically degrades with rotated text, especially when vertical and horizontal text coexist (see Supplementary Material). These issues can be mitigated by image rotation or object manipulation. The model also struggles with stylized text, such as brand logos or decorative fonts. Therefore, embodied agents' reading pipelines must account for these limitations.

\subsection{Human versus embodied agent evaluation}
\label{sec:human-vs-agent-eval}

We assigned 108 tasks, across three difficulty levels and randomized store layouts, to human participants, recording their completion time and success rate. Table~\ref{human-evaluation-results} shows easy tasks were significantly faster, as they typically involved only one item, while more items increased navigation time. Surprisingly, average tasks took longer than difficult ones, mainly due to participants' unfamiliarity with the barcode scanner and because not all difficult tasks required checkout.

\begin{table}[htbp]
\begin{center}
\caption{Performance evaluation of human versus embodied agent on the SariBench tasks based on average time to complete and completion rate. Embodied agent evaluation is limited to easy tasks. \textbf{L1/L2/L3}: Layout 1, Layout 2, Layout 3. \textbf{HAT}: Human average time in seconds. \textbf{AAT}: Embodied agent average time. \textbf{HCR}: Human completion rate. \textbf{ACR}: Embodied agent completion rate.}
\label{human-evaluation-results}
\small
\begin{tabular}{lcccc} 
\toprule
\textbf{Difficulty} & \textbf{HAT $\downarrow$} & \textbf{HCR\% $\uparrow$} & \textbf{AAT $\downarrow$} & \textbf{ACR\% $\uparrow$} \\
\hline
Easy-L1 & 47 & 88.88 & 780 & 68.63\\
Easy-L2 & 73 & 100.00 & 660 & 45.10\\
Easy-L3 & 61 & 93.33 & 420 & 33.33\\
Average-L1 & 158 & 87.50 & - &  -\\
Average-L2 & 106 &  100.00 & - & -\\
Average-L3 & 84 & 100.00 & - & -\\
Difficult-L1 & 76 & 100.00 & - & -\\
Difficult-L2 & 136 & 100.00 & - & -\\
Difficult-L3 & 113 & 100.00 & - & -\\
\bottomrule
\end{tabular}
\end{center}
\end{table}

Our findings highlight a critical disparity in easy task performance: humans consistently outperformed the embodied agent in efficiency and effectiveness. We evaluated the embodied agent's end-to-end task success by manual visual inspection: success was task completion within 45 minutes; failure was exceeding this limit or if it enters ``mode collapse'' (i.e., getting lost). The embodied agent's completion times were up to 16 times longer than humans', with success rates under 70\% compared to human rates often near 100\%. This substantial proficiency gap is partly due to the VLM's inherent computational overhead, where text generation and inference time significantly prolong the embodied agent's task completion. Despite fully utilizing the designed APIs, the embodied agent's performance fell short, indicating challenges with the VLM's overall optimal reasoning and decision-making for these embodied tasks. The embodied agent's Easy-L1 to L3 performance is a solvability proof-of-concept for Sari Sandbox, not an optimized benchmark. Harder task evaluations are future work, as our current focus is the sandbox itself. While humans generally excelled, their performance was nuanced by factors like perseverance and occasional carelessness, explaining slight dips in completion rates (e.g., Easy-L1 at 88.88\%). Notably, reported motion sickness among participants is a key consideration; though not directly impacting metrics, this VR response could affect user experience and engagement. Future research should explore mitigation strategies to enhance comfort and data reliability.

\subsection{Participant thought process flowcharts}
\usetikzlibrary{shapes.geometric, arrows.meta}

\tikzstyle{startstop} = [rectangle, rounded corners,  text centered, draw=black, fill=red!30]
\tikzstyle{process} = [rectangle, text centered, draw=black, fill=blue!30]
\tikzstyle{decision} = [rectangle, text centered, draw=black, fill=green!30]
\tikzstyle{motion} = [rectangle, text centered, draw=black, fill=orange!30]
\tikzstyle{reason} = [rectangle, text centered, draw=black, fill=purple!30]
\tikzstyle{arrow} = [thick,->,>=stealth]

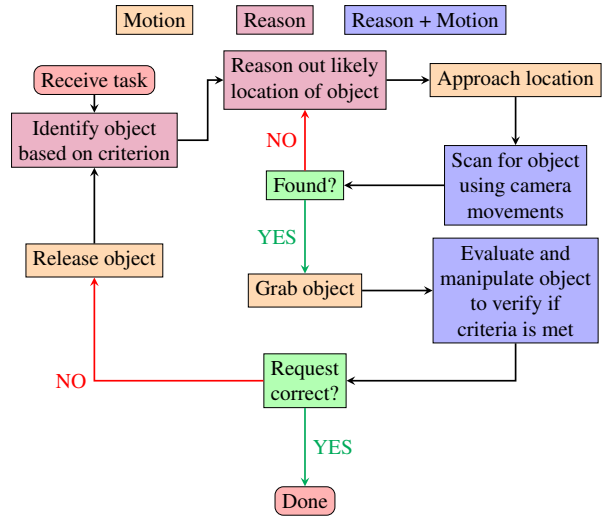
\begin{figure}[h]
\centering
\scalebox{0.8}{
\begin{tikzpicture}[node distance=1cm and 1cm, nodes = {draw,align=center}]

\node (task) [startstop] {Receive task};
\node (identify) [reason, below of=task] {Identify object\\based on criterion};
\node (reasonloc) [reason, right of=task, xshift=2.5cm] {Reason out likely\\location of object};
\node (found) [decision, below of=reasonloc, yshift=-0.75cm] {Found?};
\node (grab) [motion, below of=found, yshift=-0.75cm] {Grab object};
\node (correct) [decision, below of=grab, yshift=-0.5cm] {Request\\correct?};
\node (takeinfo) [startstop, below of=correct, yshift=-1cm] {Done};
\node (release) [motion, below of=identify, yshift=-1cm] {Release object};

\node (label-reason) [reason, above of=reasonloc, xshift=-0.5cm] {Reason};
\node (label-motion) [motion, left of=label-reason, xshift=-1cm] {Motion};
\node (label-process) [process, right of=label-reason, xshift=1.5cm] {Reason + Motion};

\node (approachloc) [motion, right of=reasonloc, xshift=2.5cm] {Approach location};
\node (scan) [process, below of=approachloc, yshift=-0.75cm] {Scan for object\\using camera\\movements};
\node (eval) [process, below of=scan, yshift=-0.75cm] {Evaluate and\\manipulate object\\to verify if\\criteria is met};

\draw [arrow] (task) -- (identify);
\draw [arrow] (identify.east) |- ++(0.5,0) |- (reasonloc.west) ;
\draw [arrow] (reasonloc) -- (approachloc);
\draw [arrow, Green!90] (found) -- node[left, draw=none] {YES} (grab);
\draw [arrow] (eval) |- (correct);
\draw [arrow, Green!90] (correct) -- node[right, draw=none] {YES} (takeinfo);

\draw [arrow, red] (found) -- node[left, draw=none] {NO} (reasonloc.south);
\draw [arrow, red] (correct) -| node[left, draw=none] {NO} (release);

\draw [arrow] (approachloc) -- (scan);
\draw [arrow] (scan.west) -- (found.east);
\draw [arrow] (grab) -- (eval);

\draw [arrow] (release.north) -- (identify.south);

\end{tikzpicture}
}

\caption{General thought process for easy tasks.}
\label{fig:general-thought-process-easy}
\end{figure}

\begin{figure}[ht]
\centering
\scalebox{0.8}{
\begin{tikzpicture}[node distance=1cm and 1cm, nodes = {draw,align=center}]

\node (task) [startstop] {Receive Task};
\node (identify) [reason, below of=task] {Identify object\\based on criterion};
\node (reasonloc) [reason, right of=task, xshift=2.5cm] {Reason out likely\\location of object};
\node (found) [decision, below of=reasonloc, yshift=-0.75cm] {Found?};
\node (grab) [motion, below of=found, yshift=-0.75cm] {Grab object};
\node (correct) [decision, below of=grab, yshift=-0.5cm] {Request\\correct?};
\node (release) [motion, below of=identify, yshift=-1cm] {Release object};

\node (label-reason) [reason, above of=reasonloc, xshift=-0.5cm] {Reason};
\node (label-motion) [motion, left of=label-reason, xshift=-1cm] {Motion};
\node (label-process) [process, right of=label-reason, xshift=1.5cm] {Reason + Motion};

\node (approachloc) [motion, right of=reasonloc, xshift=2.5cm] {Approach location};
\node (scan) [process, below of=approachloc, yshift=-0.75cm] {Scan for object\\using camera\\movements};
\node (eval) [process, below of=scan, yshift=-0.76cm] {Evaluate and\\manipulate object\\to verify if\\criteria is met};

\node (approachcheckout) [motion, below of=correct, yshift=-0.7cm] {Approach checkout};
\node (pressstart) [process, below of=approachcheckout, yshift=-0.3cm] {Press start\\button};
\node (locatebarcode) [process, below of=pressstart, yshift=-0.3cm] {Locate the\\barcode};
\node (orientbarcode) [process, below of=locatebarcode, yshift=-0.3cm] {Orient\\barcode};
\node (scanpay) [process, below of=orientbarcode, yshift=-0.3cm] {Scan and\\pay item(s)};
\node (done) [startstop, below of=scanpay] {Done};

\draw [arrow] (scanpay) -- (done);

\draw [arrow] (task) -- (identify);
\draw [arrow, Green!90] (correct) -- node[left, draw=none]{YES} (approachcheckout);
\draw [arrow] (identify.east) |- ++(0.5,0) |- (reasonloc.west) ;
\draw [arrow] (reasonloc) -- (approachloc);
\draw [arrow, Green!90] (found) -- node[left, draw=none] {YES} (grab);
\draw [arrow] (eval) |- (correct);

\draw [arrow, red] (found) -- node[left, draw=none] {NO} (reasonloc.south);
\draw [arrow, red] (correct) -| node[left, draw=none] {NO} (release);

\draw [arrow] (approachloc) -- (scan);
\draw [arrow] (scan.west) -- (found.east);
\draw [arrow] (grab) -- (eval);

\draw [arrow] (approachcheckout) -- (pressstart);
\draw [arrow] (pressstart) -- (locatebarcode);
\draw [arrow] (locatebarcode) -- (orientbarcode);
\draw [arrow] (orientbarcode) -- (scanpay);

\draw [arrow] (release.north) -- (identify.south);

\end{tikzpicture}
}

\caption{General thought process for average tasks.}
\label{fig:general-thought-process-inter}
\end{figure}
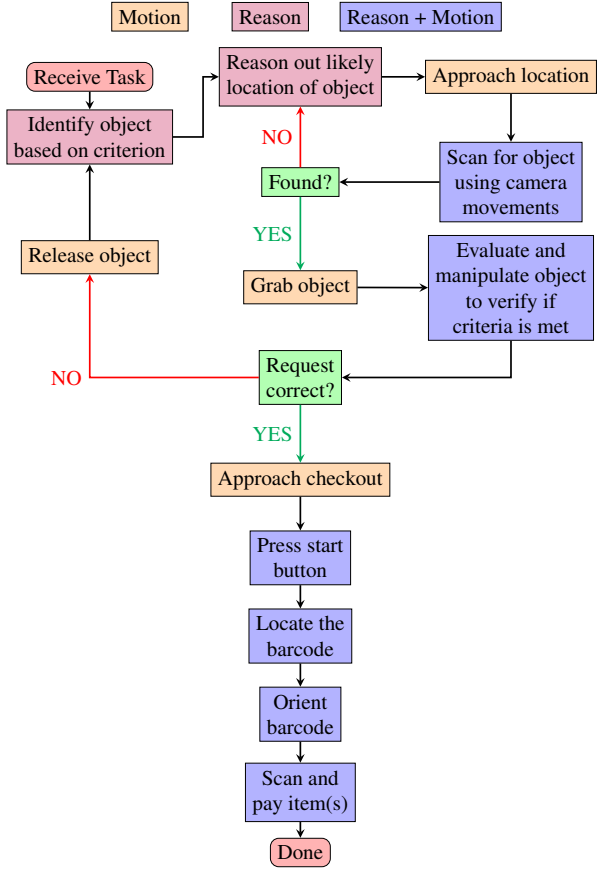

We include flowcharts illustrating the thought processes of participants during object retrieval and checkout tasks, derived from think-aloud protocols. The Easy Task Flowchart (Figure \ref{fig:general-thought-process-easy}, e.g., \textit{``Find and pick up a soda''}) involves a loop of reasoning about the object’s location, scanning, and verifying if it meets the criterion before grabbing or releasing. The Average Task Flowchart (Figure \ref{fig:general-thought-process-inter}, e.g., \textit{``Find Koko Krunch, check for artificial flavors, and scan it if none are present''}) adds interpretation and conditional actions. After verification, participants proceed to checkout, orient the barcode, and complete the scan and payment.

In the Difficult Task Flowchart (Figure ~\ref{fig:general-thought-process-diff}), participants first gather information by performing simpler sub-tasks, then apply this knowledge to answer a more complex query. Actions are categorized into \textit{Reason}, \textit{Motion}, and \textit{Reason + Motion}. Decision points often involve reassessment and retries, revealing how humans manage uncertainty and adapt strategies. Compound actions demonstrate how participants mentally bundle related physical and cognitive steps, while selectively recalling earlier information to support later goals which underscores the role of memory and flexible planning in completing complex tasks. These diagrams show how humans integrate reasoning with action, serving as useful models for hybrid agent design.

\tikzstyle{startstop} = [rectangle, rounded corners, minimum width=1cm, minimum height=0.5cm,text centered, draw=black, fill=white, font=\small]

\tikzstyle{process} = [rectangle, minimum width=1cm, minimum height=0.5cm, text centered, draw=black, fill=blue!30, font=\small]
\tikzstyle{reason} = [rectangle, minimum width=1cm, minimum height=0.5cm, text centered, draw=black, fill=yellow!50, font=\small]
\tikzstyle{motion} = [rectangle, minimum width=1cm, minimum height=0.5cm, text centered, draw=black, fill=red!30, font=\small]
\tikzstyle{decision} = [diamond, minimum width=1cm, minimum height=0.5cm, text centered, draw=black, fill=yellow!50, font=\small]
\tikzstyle{arrow} = [thick,->,>=stealth]
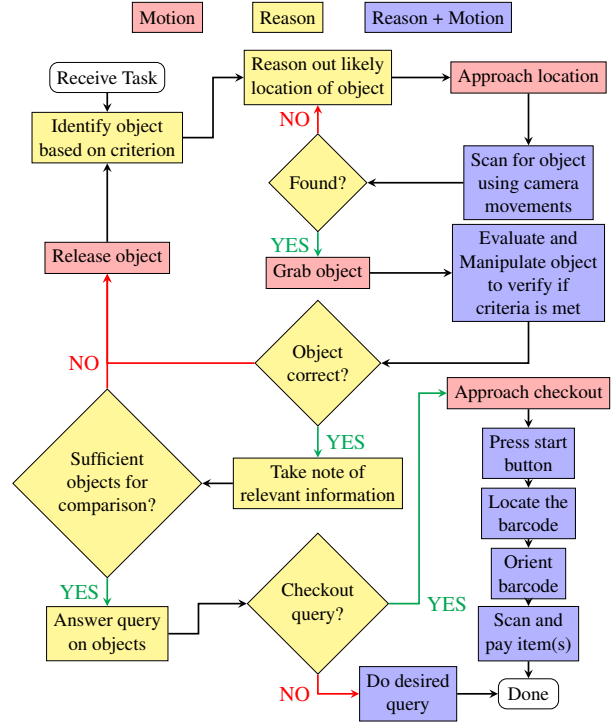
\begin{figure}[ht]
\centering
\scalebox{0.8}{
\begin{tikzpicture}[node distance=1cm and 1cm, nodes = {draw,align=center}]

\node (task) [startstop] {Receive Task};
\node (identify) [reason, below of=task] {Identify object\\based on criterion};
\node (reasonloc) [reason, right of=task, xshift=2.5cm] {Reason out likely \\location of object};
\node (found) [decision, below of=reasonloc, yshift=-0.75cm] {Found?};
\node (grab) [motion, below of=found, yshift=-0.5cm] {Grab object};
\node (correct) [decision, below of=grab, yshift=-0.5cm] {Object\\ correct?};
\node (takeinfo) [reason, below of=correct, yshift=-1cm] {Take note of\\relevant information};
\node (sufficient) [decision, left of=takeinfo, xshift=-2.5cm] {Sufficient\\objects for\\comparison?};
\node (release) [motion, below of=identify, yshift=-1cm] {Release object};

\node (label-reason) [reason, above of=reasonloc, xshift=-0.5cm] {Reason};
\node (label-motion) [motion, left of=label-reason, xshift=-1cm] {Motion};
\node (label-process) [process, right of=label-reason, xshift=1.5cm] {Reason + Motion};

\node (approachloc) [motion, right of=reasonloc, xshift=2.5cm] {Approach location};
\node (scan) [process, below of=approachloc, yshift=-0.75cm] {Scan for object\\using camera\\movements};
\node (eval) [process, below of=scan, yshift=-0.50cm] {Evaluate and\\ Manipulate object\\to verify if\\criteria is met};

\node (answerquery) [reason, below of=sufficient, yshift=-1.5cm] {Answer query\\on objects};
\node (checkoutquery) [decision, below of=takeinfo, yshift=-1cm] {Checkout\\query?};
\node (approachcheckout) [motion, below of=eval, yshift=-1cm] {Approach checkout};
\node (pressstart) [process, below of=approachcheckout] {Press start\\ button};
\node (locatebarcode) [process, below of=pressstart] {Locate the\\barcode};
\node (orientbarcode) [process, below of=locatebarcode] {Orient\\barcode};
\node (scanpay) [process, below of=orientbarcode] {Scan and\\pay item(s)};
\node (done) [startstop, below of=scanpay] {Done};

\node (desiredquery) [process, left of=done, xshift=-1cm] {Do desired\\query};

\draw [arrow] (scanpay) -- (done);

\draw [arrow] (task) -- (identify);
\draw [arrow] (identify.east) |- ++(0.5,0) |- (reasonloc.west) ;
\draw [arrow] (reasonloc) -- (approachloc);
\draw [arrow, Green!90] (found) -- node[left, draw=none] {YES} (grab);
\draw [arrow] (eval) |- (correct);
\draw [arrow, Green!90] (correct) -- node[right, draw=none] {YES} (takeinfo);
\draw [arrow] (takeinfo) -- (sufficient);
\draw [arrow, Green] (sufficient) -- node[left, draw=none] {YES} (answerquery);

\draw [arrow, red] (found) -- node[left, draw=none] {NO} (reasonloc.south);
\draw [arrow, red] (correct) -| node[left, draw=none] {NO} (release);
\draw [arrow, red] (sufficient.north) -- (release.south);

\draw [arrow] (approachloc) -- (scan);
\draw [arrow] (scan.west) -- (found.east);
\draw [arrow] (grab) -- (eval);

\draw [arrow] (answerquery.east) |- ++(0.5,0) |- (checkoutquery.west);
\draw [arrow, red] (checkoutquery) |- node[left, draw=none] {NO} (desiredquery);
\draw [arrow, Green] (checkoutquery.east) |- ++(0.5,0) node[right, draw=none] {YES} |- (approachcheckout);
\draw [arrow] (approachcheckout) -- (pressstart);
\draw [arrow] (pressstart) -- (locatebarcode);
\draw [arrow] (locatebarcode) -- (orientbarcode);
\draw [arrow] (orientbarcode) -- (scanpay);
\draw [arrow] (desiredquery) -- (done);

\draw [arrow] (release.north) -- (identify.south);

\end{tikzpicture}
}

\caption{General thought process for difficult tasks.}
\label{fig:general-thought-process-diff}
\end{figure}
\section{Conclusion and future work}

We introduce Sari Sandbox, a synthetic grocery environment with 250 items for training embodied agents, and its accompanying SariBench dataset, both serving as our prime contributions. Sari Sandbox provides an API-driven action set for varied task difficulties and uniquely incorporates human thought processes for benchmarking via SariBench's captured tasks and human demonstrations. Future efforts will focus on optimizing performance to a stable 60 FPS on mid-range desktops, expanding the dataset with dynamic additions, and enhancing realism through deformable objects and mesh colliders. A critical focus is broadening embodied agent evaluation across all SariBench difficulties to design more sophisticated embodied agents with improved navigation, perception, and manipulation, bridging the human-agent performance gap. This includes dedicated research into optimal VLM context engineering to enhance reasoning and planning. We also plan to streamline dataset generation via automated annotations, update store designs, and develop a user-friendly scene creation tool for rapid prototyping and diverse real-world simulations.
{
    \small
    \bibliographystyle{IEEEtran}
    \bibliography{main}
}

\clearpage
\setcounter{page}{1}
\renewcommand{\thesection}{S\arabic{section}}
\setcounter{section}{0}  
\renewcommand{\thefigure}{S\arabic{figure}}
\renewcommand{\thetable}{S\arabic{table}}
\setcounter{figure}{0}
\setcounter{table}{0}

\maketitlesupplementary

\section{Texture fidelity}
\label{supp_sec: texture_fidelity_supp}
\begin{figure}[htbp]
    \centering
    \includegraphics[width=1\linewidth]{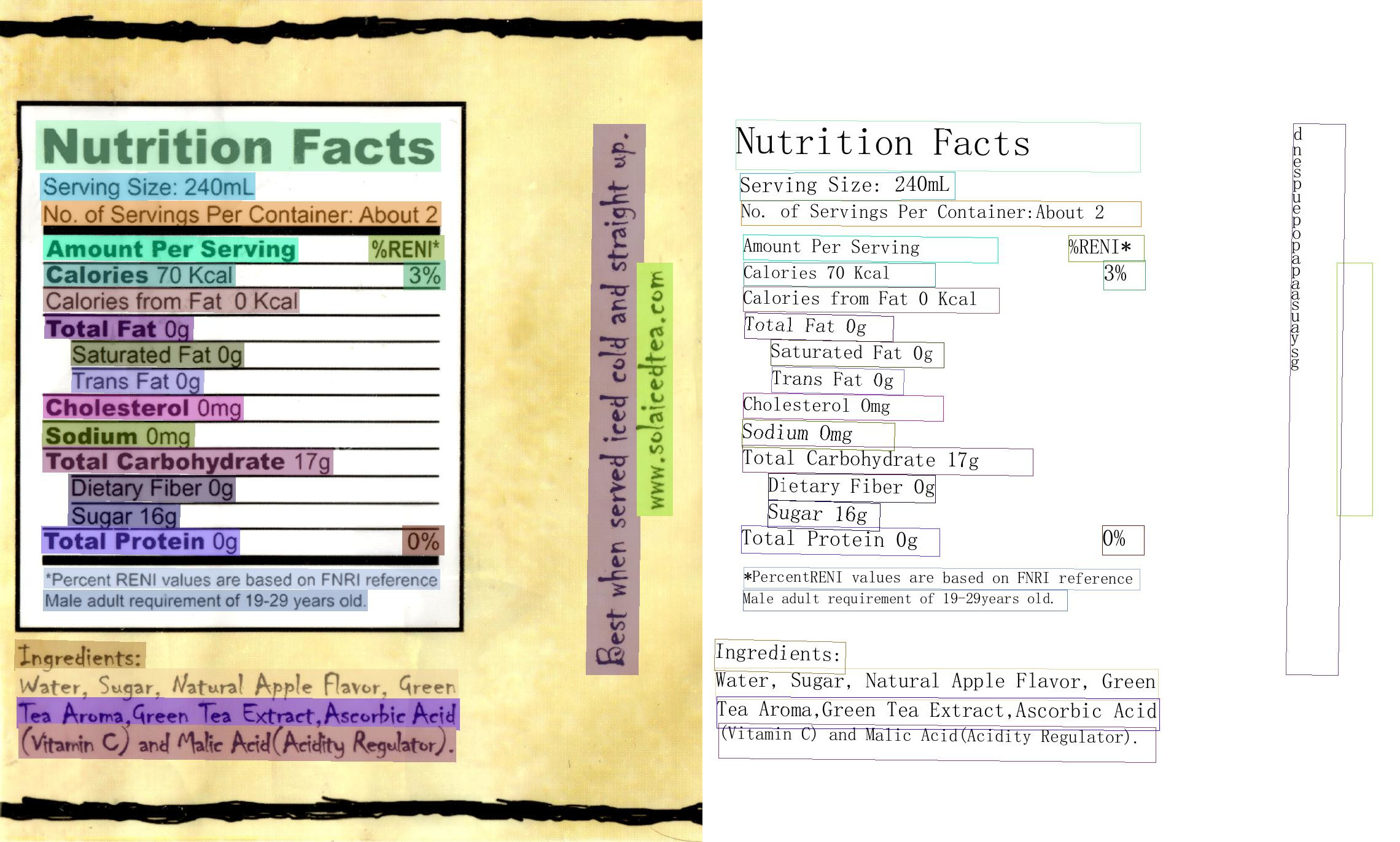}
    \caption{PaddleOCR failed in identifying the rotated text because majority is horizontal text.}
    \label{fig:paddle1}
\end{figure}

\begin{figure}[htbp]
    \centering
    \includegraphics[width=1\linewidth]{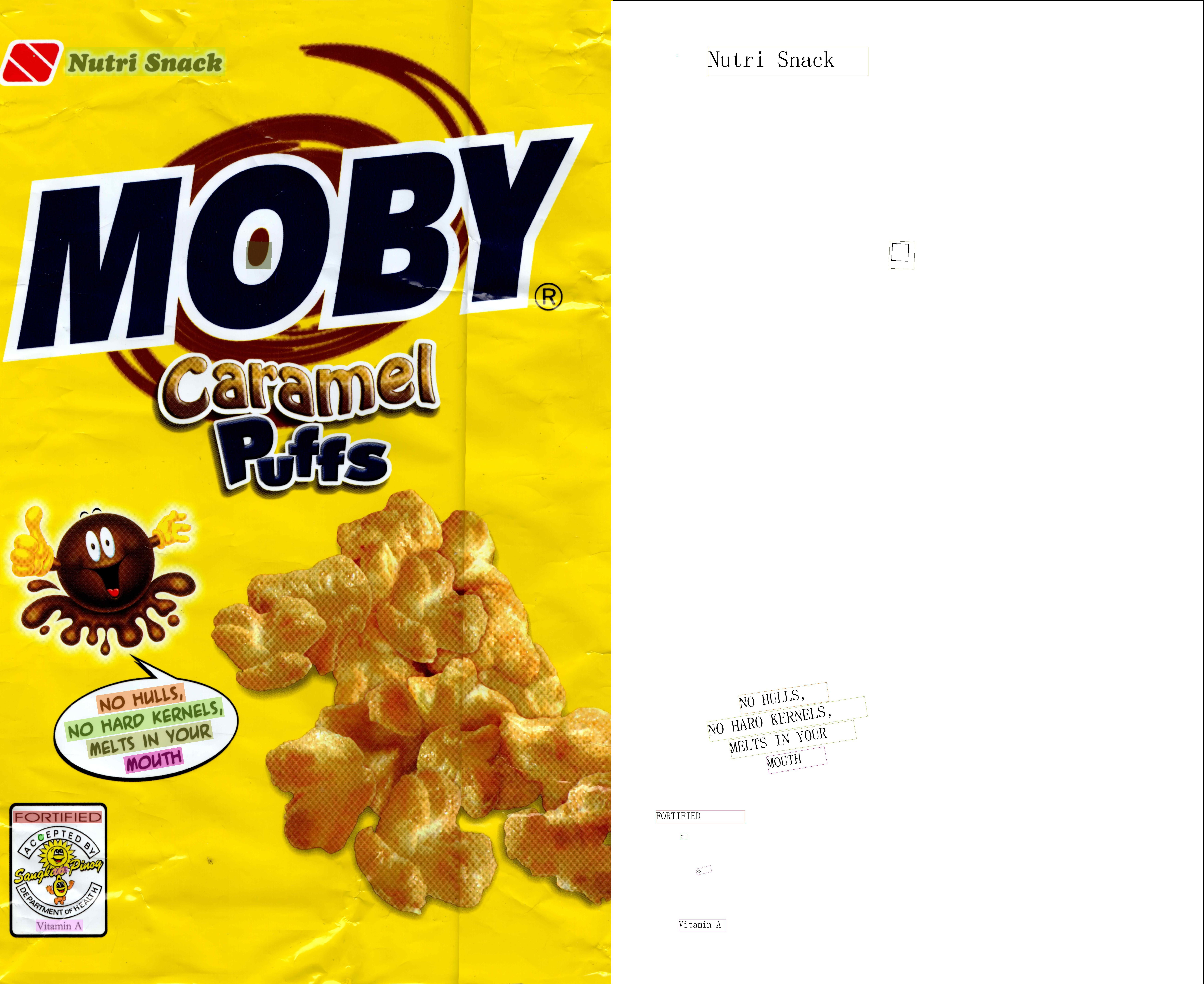}
    \caption{PaddleOCR failed in identifying stylized text of the Moby Brand despite being able to identify texts in the same image.}
    \label{fig:paddle2}
\end{figure}

As discussed in \cref{sec:texture-fidelity}, we evaluated texture fidelity using PaddleOCR~\cite{paddleocr2020}. While it performs well on structured product labels, it exhibits limitations in more challenging scenarios. Figure~\ref{fig:paddle1} and Figure~\ref{fig:paddle2} provide qualitative examples where OCR accuracy declines. Figure~\ref{fig:paddle1} presents a case involving mixed horizontal and vertical text orientations, which often result in incorrect segmentation or recognition. Figure~\ref{fig:paddle2} highlights the challenges posed by stylized fonts and logo-like text, where decorative design elements interfere with accurate character detection. These examples encourages OCR-aware scene design and the potential need for active viewpoint control in embodied agent pipelines.

\section{Embodied agent}
\label{supp_sec: embodied_agent}

\begin{figure}[htbp]
\centering
\resizebox{0.70\columnwidth}{!}{%
\begin{circuitikz}
\tikzstyle{every node}=[font=\LARGE]
\draw [ fill={rgb,255:red,222; green,222; blue,222} , line width=1.1pt , rounded corners = 22.5] (1.5,12.25) rectangle (5,10.75);
\node [font=\Huge] at (3.25,11.5) {\textbf{Human}};
\draw [ fill={rgb,255:red,255; green,225; blue,173} , line width=1.1pt ] (11.75,12.5) rectangle  node {\Huge \textbf{Sandbox}} (15.5,10.75);
\draw [ fill={rgb,255:red,173; green,209; blue,255} , line width=1.1pt ] (6.75,12.5) rectangle  node {\Huge \textbf{Agent}} (9.5,10.75);
\draw [line width=1pt, ->, >=Stealth] (9,12.5) .. controls (10,13.75) and (13.25,13.25) .. (13.75,12.5) ;
\node [font=\Huge] at (11.25,13.75) {\textbf{Action}};
\draw [line width=1pt, ->, >=Stealth] (13.75,10.75) .. controls (13,9.5) and (9.5,9.75) .. (9,10.75) ;
\node [font=\Huge] at (11.5,9.25) {\textbf{Feedback}};
\draw [ fill={rgb,255:red,222; green,222; blue,222} , line width=1.1pt , rounded corners = 22.5] (6.25,8.75) rectangle (9.75,7.25);
\node [font=\Huge] at (8,8) {\textbf{STOP}};
\draw [->, >=Stealth, dashed] (8,10.75) -- (8,8.75);
\draw [<->, >=Stealth, dashed] (5,11.5) -- (6.75,11.5);
\end{circuitikz}
}%
\caption{The elementary concept of an autonomous agent.}
\label{fig:autonomous_agent_diagram}
\end{figure}
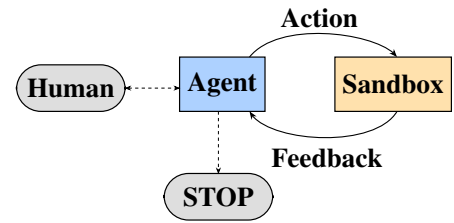

An agent is an entity that perceives its environment through sensors and acts upon it using actuators~\cite{RusselNorvig_AIMA_2016}. As illustrated in Figure~\ref{fig:autonomous_agent_diagram}, the agent, a modular system, operates within its environment (sandbox) through a continuous loop of perception, cognition, and action to achieve a set goal. Upon goal completion, it halts until a human provides a new directive. This paradigm is powerful as it shifts from static programming to a dynamic, interactive model where systems are both observers and actors.

\begin{figure*}[htbp]
    \centering
    \includegraphics[trim={0cm 2.05cm 0cm 2.05cm}, clip, width=\linewidth]{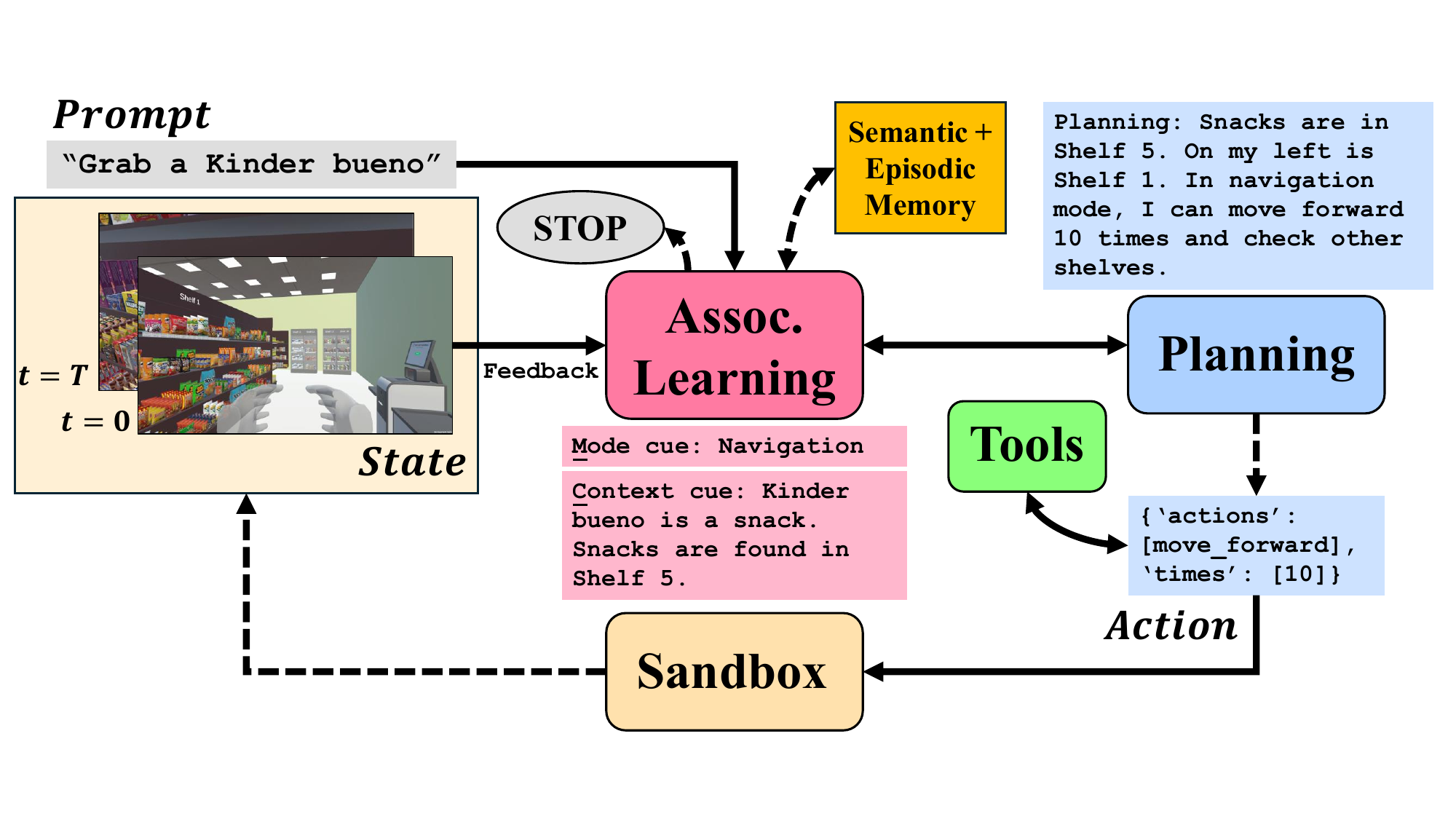}
    \caption{An overview of our agentic pattern. The process begins with a prompt, initiating a continuous loop. In the initial \textbf{associative learning} step, the embodied agent consults its semantic and episodic memory, and synthesizes its current state and the given task into \underline{m}ode and \underline{c}ontext (M+C) cues. These cues are included as inputs for the \textbf{planning} step, where the VLM generates an action sequence. This sequence, comprising elementary actions and potential tool calls, is executed within the sandbox. The loop then repeats with the new state, continuing until the associative learning step issues a \texttt{STOP} command.}
    \label{fig:embodied_agent_rich_graphic}
\end{figure*}

For our agent's cognitive engine, we required a VLM instead of an LLM, incorporating visual sensory inputs. We based our selection on the Massive Multi-discipline Multimodal Understanding and Reasoning (MMMU)~\cite{yue2024mmmu}. At the time of evaluation, Gemini 2.5 Pro was state-of-the-art, achieving a score of 84.0 on the MMMU validation set (using its experimental Deep Think mode), nearing the human expert benchmark of 88.6~\cite{google_gemini_2025}.

In our sandbox, the agent faces two core challenges: autonomous navigation and object manipulation, both vital for product search and retrieval. We adapted the ReAct framework for our needs. To focus on core capabilities, we simplified the embodied agent's navigation and manipulation; complex behaviors like multi-item handling or checkout were excluded. We forego A* planning~\cite{hart1968formal} as our embodied agent lacks access to a pre-built grid-map, relying solely on visual input during navigation. Our primary objective is not to develop a state-of-the-art agent, but rather to construct a functional one that rigorously tests the practicality and usability of the APIs designed in Section~\ref{subsection:api} through basic item search and retrieval tasks.

Our embodied agent operates using the pattern in Figure~\ref{fig:embodied_agent_rich_graphic}, which is designed for both efficiency and control when processing a grocery task (e.g., \textit{``Find a healthy snack''}). Our key approach here is that the embodied agent generates an action sequence---a chunk of related commands---rather than a single action per  cycle. This approach is highly effective for repetitive tasks such as grocery shopping (e.g., \textit{``navigate to Shelf 1, search; navigate to Shelf 2, search; navigate to Shelf 3, grasp''}). This pattern effectively manages such decomposed, sequential steps. Furthermore, this design substantially offloads the cognitive load from the VLM. The associative learning step pre-processes and provides mode cues (e.g., indicating navigation or manipulation) and context cues. This means the planning step does not need to deliberate on these operational states, streamlining its decision-making. Consequently, this pattern makes debugging easier and provides more robust control over the outputs.

The agentic loop operates through three key steps. First, the \textbf{associative learning step} queries the sandbox for the embodied agent's current state (coordinates, visual input) at each step. It then processes this state against the prompt to generate two cues for the next stage: a \textbf{mode cue} and a \textbf{context cue}. The mode cue switches the embodied agent's operational state between navigation and manipulation, constraining the VLM to specific symbolic actions for improved reliability. The context cue provides recall information from the embodied agent's semantic and episodic memory. Second, \textbf{planning step} takes this state information, along with the mode cue, context cue, and available tools (function-calling APIs), and passes them to the VLM, which generates a structured action sequence. Finally, during the \textbf{execution step}, an external parser translates this sequence into executable API calls dispatched to the sandbox, yielding an observable result. The embodied agent perceives this new state, and the loop repeats until the associative learning step issues a \texttt{STOP} command.

To overcome the stateless nature inherent in VLM-powered embodied agents, our embodied agent explicitly models memory, leveraging the four-part cognitive architecture. This framework, managed within our agentic pattern, comprises the following components: \textbf{procedural memory} which contains the embodied agent's core rules and skills, implemented as the system instructions provided to the VLM; \textbf{working memory} which contains the immediate interaction history, managed by caching and injecting context within the VLM's context window; \textbf{semantic memory} which contains the factual knowledge about the environment; and \textbf{episodic memory} which contains the distilled takeaways from the embodied agent's own experiences. To implement the semantic memory, we first created a base semantic memory which is a text file containing the store layout, product locations (e.g., \textit{``Shelf 1 contains cereals''}), and rudimentary directions from the embodied agent's spawn point to any shelf (mimicking natural instructions). Similarly, the episodic memory is implemented as a text file, which starts blank at the beginning of each task.

The semantic and episodic memories are managed by a memory writing operation that occurs during the associative learning step. At each timestep, this module consults both memory files. It uses the semantic memory to ground the embodied agent in its environment. After action execution, the associative learning step updates the episodic memory by synthesizing a three-point reflection on the action that had just been executed: a dense summary of what occurred, what actions worked, and what to avoid in the future. The key information recalled from both memories is then encoded into the context and mode cues, and passed to the planning step to reduce the cognitive load and enhance context. It is important to note that while the associative learning and planning steps use the same Gemini model release version, they function as distinct modules with different system instructions.

\section{Actions and tools}

Our embodied agent interacts with the Sari Sandbox environment through a defined set of actions, categorized into distinct operational modes: navigation and manipulation~(Table \ref{tab:actions-table}). These actions enable precise control over the embodied agent's movement and interaction with objects within the simulated grocery store.

The navigation mode allows the agent to control its body's position and orientation. This includes fundamental actions such as \texttt{move\_forward}, which advances the agent by 0.1 units, and \texttt{pan\_left} and \texttt{pan\_right}, which rotate the agent's view horizontally by 2.5-degree increments. While these appear as single, high-level API calls, their underlying implementation involves iterative, atomic calls to the simulator's core API functions. For instance, \texttt{move\_forward} directly invokes \texttt{TransformAgent((0, 0, 0.1), (0, 0, 0))}, allowing for controlled, fine-grained movement up to a predefined unit limit. Similarly, \texttt{pan\_left} and \texttt{pan\_right} are built upon \texttt{TransformAgent((0, 0, 0), (0, -2.5, 0))} and \texttt{TransformAgent((0, 0, 0), (0, +2.5, 0))}, respectively, to control the agent's yaw rotation.

\begin{table}[!ht]

\centering
\caption{Different modes of operation and their associated actions. Action descriptions: \texttt{move\_forward} to move the embodied agent forward by 0.1 units in the sandbox. \texttt{pan\_left} and \texttt{pan\_right} to pan left and right by 2.5 degrees, respectively. \texttt{center\_object\_on\_screen} to center the embodied agent's body on the target object in the frame. \texttt{retrieve\_item} to approach the target object, grab it with the embodied agent's hand, and inspect it. Navigation and manipulation invokves \texttt{TransformAgent} and \texttt{TransformHands}, respectively.}
\label{tab:actions-table}
\small
\begin{tabular}{lp{5cm}} 
\toprule
\textbf{Mode} & \textbf{Actions} \\
\midrule
Navigation & \texttt{move\_forward}, \texttt{pan\_left}, \texttt{pan\_right} \\
Manipulation & \texttt{center\_object\_on\_screen}, \texttt{retrieve\_item} \\
\bottomrule
\end{tabular}

\end{table}

The manipulation mode enables the agent to interact directly with items in the environment. This includes actions like \texttt{center\_object\_on\_screen} and \texttt{retrieve\_item}.

The \texttt{center\_object\_on\_screen} action leverages Gemini 2.5 Pro for object detection. It uses visual inputs (obtained via first-person point-of-view screenshot within the sandbox) to calculate the target object's bounding box. The VLM's perception output after calling \texttt{loc\_object} tool, providing \texttt{ymin}, \texttt{xmin}, \texttt{ymax}, \texttt{xmax} coordinates, is then translated into pixel coordinates. Based on the object's horizontal and vertical deviation from the screen center, the agent directly invokes \texttt{TransformAgent} to perform precise yaw and pitch rotations, aligning its perspective with the object.

The \texttt{retrieve\_item} action is a more complex, compound behavior that orchestrates several steps:
\begin{itemize}
    \item \textbf{Depth estimation}. The agent first moves generally towards the detected target using visual input and estimated depth via \texttt{est\_depth} tool. This often involves \texttt{move\_forward} actions, which, as described, translate to repeated \texttt{TransformAgent((0, 0, 0.1), (0, 0, 0))} calls.
    \item \textbf{Fine-tuning orientation}. It then adjusts its orientation to face a cardinal direction for consistent alignment, again utilizing \texttt{TransformAgent} for precise yaw control.
    \item \textbf{Horizontal centering}. The embodied agent performs a \texttt{strafe\_to\_center} operation to precisely align itself with the object. This action calculates the horizontal offset of the target object's bounding box from the image center. It then converts this pixel offset into a required linear movement in world units. Subsequently, \texttt{strafe\_to\_center} executes a series of granular calls which correspond to \texttt{TransformAgent((+0.1, 0, 0), (0, 0, 0))} or \texttt{TransformAgent((-0.1, 0, 0), (0, 0, 0))} to incrementally shift the embodied agent's body sideways until the object is horizontally centered in its view.
    \item \textbf{Final approach and interaction}. The embodied agent moves to the item's immediate vicinity and executes the physical \texttt{grab\_and\_read} operation. This critical step involves a choreographed series of atomic hand movements directly implemented via the \texttt{TransformHands} API. Specifically, the agent can extend its hands forward by adjusting their Z-axis position, pull them backward for Z-axis retraction, or raise and lower them to control their Y-axis position. Rotational actions, also achieved by \texttt{TransformHands}, allow for precise manipulation of the hand's yaw rotation. Once positioned, the agent can grasp the item using \texttt{ToggleLeftGrip} API calls to simulate a grip. Following the grab, the \texttt{grab\_and\_read\_item} operation initiates a new screenshot and then processes this image using an Optical Character Recognition (OCR) tool via \texttt{ocr\_object} to extract any text present on the item, thereby simulating visual inspection. These primitive hand and vision-based interactions are crucial for the embodied agent to accurately reach, grasp, and inspect the target item, even though they are not exposed as high-level API actions in the table.
\end{itemize}

The VLM serves as the cognitive engine for these actions. It processes visual information and textual prompts to determine the appropriate sequence of actions and their parameters. Tools found in Table~\ref{tab:tools-table} are integral to the embodied agent's perception and decision-making loop, particularly for tasks requiring object identification and precise interaction. The structured nature of these higher-level actions, built upon the fundamental \texttt{TransformAgent} and \texttt{TransformHands} APIs, ensures the embodied agent can perform complex grocery tasks by decomposing them into manageable, executable steps.

\begin{table}[!htbp]

\centering
\caption{Tools that are part of an elementary action and their corresponding purposes.}
\small
\begin{tabular}{lp{5cm}} 
\toprule
\textbf{Tool} & \textbf{Purpose} \\
\midrule
\texttt{loc\_object} & Uses Gemini 2.5 Pro's object localization capability to output bounding box coordinates of a specified item or item of interest. Format: \texttt{[ymin, xmin, ymax, xmax]}. Part of: \texttt{center\_object\_on\_screen}.\\
\texttt{ocr\_object} & Uses PaddleOCR~\cite{paddleocr2020} for item inspection via optical character recognition (OCR). Part of: \texttt{retrieve\_item}.\\
\texttt{est\_depth} & Uses Depth-Anything-V2~\cite{yang2024depth} Small for computing the distance between target object and embodied agent before item inspection. Part of: \texttt{retrieve\_item}.\\
\bottomrule
\end{tabular}
\label{tab:tools-table}
\end{table}

\end{document}